\documentclass{article}

\usepackage{spconf}
\usepackage{amsmath}
\usepackage{caption}
\usepackage{float}
\usepackage{graphicx}
\usepackage{paralist}
\usepackage{color}

\floatstyle{ruled}
\newfloat{algorithm}{!b}{loa}
\floatname{algorithm}{Algorithm}
\def\Vec#1{{\boldsymbol{#1}}}
\def\Mat#1{{\boldsymbol{#1}}}

\def\eqn#1{\mbox{Eqn.\hspace{0.5ex}(\ref{#1})}}

\begin{document}

\title
  {
  K-Tangent Spaces on Riemannian Manifolds\\
  for Improved Pedestrian Detection
  }

\name
  {
  Andres Sanin, Conrad Sanderson, Mehrtash T. Harandi, Brian C. Lovell
  }

\address
  {
  NICTA, PO Box 6020, St Lucia, QLD 4067, Australia%
  \thanks
    {
    {\bf Acknowledgements:}
    NICTA is funded by the Australian Government as represented by the {\it Department of Broadband, Communications and the Digital Economy},
    as well as the Australian Research Council through the {\it ICT Centre of Excellence} program.
    }
  \\
  University of Queensland, School of ITEE, QLD 4072, Australia
  }

\maketitle
\renewcommand{\baselinestretch}{0.95}\small\normalsize

\begin{abstract}

\noindent
For covariance-based image descriptors,
taking into account the curvature of the corresponding feature space
has been shown to improve discrimination performance.
This is often done through representing the descriptors as points on Riemannian manifolds,
with the discrimination accomplished on a tangent space.
However, such treatment is restrictive as distances between arbitrary points on the tangent space
do not represent true geodesic distances, 
and hence do not represent the manifold structure accurately.
In this paper we propose a general discriminative model based on the combination of several tangent spaces,
in order to preserve more details of the structure.
The model can be used as a weak learner in a boosting-based pedestrian detection framework.
Experiments on the challenging INRIA and DaimlerChrysler datasets
show that the proposed model leads to considerably higher performance
than methods based on histograms of oriented gradients as well as previous Riemannian-based techniques.

\end{abstract}

\vspace{-0.25ex}

\section{Introduction}
\label{sec:introduction}
\vspace{-1.5ex}

Covariance descriptors have recently become popular features for pedestrian detection,
due to their relative low sensitivity to translations and scale variations,
as well as being quick to compute through integral image representations
\mbox{\cite{GualdiEtAl2011,PaisitkriangkraiEtAl2008,TuzelEtAl2008}}.
Covariance descriptors belong to the group of positive definite symmetric matrices ({\small $Sym_d^+$}),
which can be interpreted as points on Riemannian manifolds~\cite{Harandi_ECCV_2012}.
Intuitively, we can think of a manifold as a continuous surface lying in a higher dimensional Euclidean space.
By exploiting the curvature of the feature space formed by such matrices,
higher performance can be achieved than vector space approaches
such as histograms of oriented gradients (HOGs), 
previously considered as the state of the art features for pedestrian detection~\cite{DollarEtAl2011}.

While the geometry of Riemannian manifolds has been extensively studied,
with well defined metrics and operations for {\small $Sym_d^+$},
classification directly on manifolds
(eg.~dividing a manifold into two distinct parts)
is still a challenging problem~\cite{HarandiEtAl2012,Shirazi_ICIP_2012}.
Current approaches typically first map the points on the manifold to a tangent space,
where traditional machine learning techniques can be used for classification~\mbox{\cite{GuoEtAl2010,TuzelEtAl2008}}.
A~tangent space is an Euclidean space relative to a point in the manifold (the tangent pole).
Processing a manifold through a single tangent space is necessarily restrictive,
as only distances to the tangent pole are true geodesic distances.
Distances between arbitrary points on the tangent space do not represent true geodesic distances,
and hence do not represent the manifold structure accurately.

As a partial workaround to the above limitation,
Tuzel et al.~\cite{TuzelEtAl2008} recently proposed a pedestrian detection method
based on a LogitBoost cascade~\cite{FriedmanEtAl2000} over Riemannian manifolds.
Linear regression was used as a weak classifier within the cascade,
after mapping the selected training points to a tangent space that occurs at the mean of the points.
In other words, for each weak classifier a new tangent pole is obtained closer to the most critical samples.
As a consequence of boosting, the final inference on the manifold was made through multiple tangent spaces.
However, there is an inherent limitation
in that each boosting iteration has to deal with two problems at the same time:
finding the best tangent representation and finding the best discriminative model.

In this paper, 
we propose a general discriminative model based on the combination of several tangent spaces,
in order to preserve more details of the manifold structure. 
We shall refer to this model as {\it k}-tangent spaces.
The model can be used as a weak classifier in a boosting-based framework,
allowing the learning algorithm to focus on better classification
rather than finding the best representation of the manifold.
We demonstrate empirically that the proposed model leads to considerably improved pedestrian detection performance.

We continue the paper as follows.
In Section~\ref{sec:background} we overview relevant manifold operations
for the space of symmetric positive definite matrices.
The proposed model is described in Section~\ref{sec:model}.
Results from pedestrian detection experiments on two challenging datasets are given in Section~\ref{sec:experiments}.
The main findings and possible future directions are summarised in Section~\ref{sec:conclusions}.
\vspace{-1ex}
\section{Riemannian Geometry}
\label{sec:background}
\vspace{-1.5ex}

In this section we briefly review Riemannian geometry,
with a focus on the space of symmetric positive definite matrices.
Formally, a manifold is a topological space which is locally similar to an Euclidean space~\cite{Harandi_ECCV_2012,TuzelEtAl2008}.
The minimum length curve connecting two points on the manifold is called the geodesic,
and the distance between two points {\small $\Mat{X}$} and {\small $\Mat{Y}$} is given by the length of this curve.

The tangent space, {\small $T_\Mat{X}$} at {\small $\Mat{X}$},
is the plane tangent to the surface of the manifold at that point.
Geodesics (on the manifold) are related to the vectors in the tangent space.
Two operators, namely the exponential map {\small $\exp_{\Mat{X}}$}
and the logarithm map {\small $\log_{\Mat{X}}\mbox{=}\exp^{-1}_{\Mat{X}}$},
switch between manifold and tangent space at~{\small $\Mat{X}$}.
%
%
For~{\small $Sym_d^+$} the exponential and logarithm maps are defined as:

\vspace{-2ex}
\begin{small}
\begin{eqnarray}
    \exp_{\Mat{X}}\left( \Vec{y} \right) & = & \Mat{X}^{\frac{1}{2}}\exp \left(\Mat{X}^{-\frac{1}{2}} \Vec{y}\Mat{X}^{-\frac{1}{2}} \right) \Mat{X}^{\frac{1}{2}}
    \label{eqn:sym_exp_map}
    \\
    \log_{\Mat{X}}\left( \Mat{Y} \right) & = & \Mat{X}^{\frac{1}{2}}\log \left(\Mat{X}^{-\frac{1}{2}} \Mat{Y}\Mat{X}^{-\frac{1}{2}} \right) \Mat{X}^{\frac{1}{2}}
    \label{eqn:sym_log_map}
\end{eqnarray}%
\end{small}%

\noindent
where {\small $\Vec{y}$} (on the tangent space) is a representative of {\small $\Mat{Y}$} (on the manifold).
Furthermore,
{\small $\exp\left(\cdot\right)$} and {\small $\log\left(\cdot\right)$}
are matrix exponential and logarithm operators, respectively.
For symmetric positive definite matrices they can be computed through Singular Value Decomposition (SVD).
More specifically, let
\mbox{\small $\Mat{X}=\Mat{U}\Sigma\Mat{U}^T$} be the SVD of the symmetric matrix $\Mat{X}$,
then

\vspace{-2ex}
\begin{small}
\begin{eqnarray}
    \exp \left( \Mat{X} \right) & = & \Mat{U}\exp\left(\Sigma\right)\Mat{U}^T
    \label{eqn:matrix_exp}
    \\
    \log \left( \Mat{X} \right) & = & \Mat{U}\log\left(\Sigma\right)\Mat{U}^T
    \label{eqn:matrix_log}
\end{eqnarray}%
\end{small}%

\noindent
where
{\small $\exp\left(\Sigma\right)$} and {\small $\log\left(\Sigma\right)$}
are diagonal matrices with diagonal elements that are respectively equivalent to
the exponential or logarithms of the diagonal elements of matrix {\small $\Sigma$}.

Using the tangent space defined at {\small $\Mat{X}$},
the distance between {\small $\Mat{X}$} and {\small $\Mat{Y}$} can be calculated via:

\vspace{-1ex}
\begin{small}
\begin{equation}
  d^2(\Mat{X},\Mat{Y}) = \operatorname{trace}\left\{ \log^2 \left( \Mat{X}^{-\frac{1}{2}} \Mat{Y} \Mat{X}^{-\frac{1}{2}} \right) \right\}
  \label{eq:geodesic}
\end{equation}%
\end{small}%

Given a set {\small $\{ \Mat{X}_i\}_{i=1}^{N}$} of points on a manifold {\small $\mathcal{M}$},
the Karcher mean~\cite{Karcher1977} is the point on {\small $\mathcal{M}$}
that minimises the sum of distances:

\vspace{-2ex}
\begin{small}
\begin{equation}
  \Mat{\mu} = \operatorname*{arg\,min}_{X \in \mathcal{M}} \sum\nolimits_{i=1}^N{d^2(\Mat{X}_i,\Mat{X})}
  \label{eq:mean_def}
\end{equation}%
\end{small}%

\noindent
For {\small $Sym_d^+$}, the Karcher mean can be iteratively found using:

\vspace{-1ex}
\begin{small}
\begin{equation}
  \Mat{\mu}^{[t+1]} = \exp_{\Mat{\mu}^{[t]}} \left\{ N^{-1} \sum\nolimits_{i=1}^N \log_{\Mat{\mu}^{[t]}}\left( \Mat{X}_i \right) \right\} 
  \label{eq:mean_calc}
\end{equation}%
\end{small}%

%
%
%
%
\section{\small K-Tangent Spaces for Pedestrian Detection}
\label{sec:model}
\vspace{-1ex}

Traditional tangent-based discrimination approaches generally have three steps:
{\bf (i)}~select the tangent pole {\small $\Mat{X} \in \mathcal{M}$};
{\bf (ii)}~project the training samples onto the tangent space {\small $T_\Mat{X}$};
and
{\bf (iii)}~train a discriminative model $g$ (eg.~the regression function in~\cite{TuzelEtAl2008}) using the projected samples.
The discrimination performance hence depends on two characteristics:
{\bf (a)}~how well model $g$ is able to fit the data,
and
{\bf (b)}~how well the tangent representation preserves the distances between the points on the manifold.
Intuitively speaking, the larger the area covered by the tangent space,
the worse the representation.

To address the above issue, we propose the combination of multiple tangent spaces
into a single discriminative model:

\vspace{-1ex}
\begin{small}
\begin{equation}
  \lambda = \left\{ (\Mat{\mu}_k, n_{_k}, g_{_k}) \right\}_{k=1}^{K}
\end{equation}%
\end{small}%

\noindent
where,
for the {\small $k$}-th entry,
{\small $\Mat{\mu}_k$} is the tangent pole (center),
{\small $n_{_k}$} is the number of samples associated with the center,
and
{\small $g_{_k}$} is a discriminative model associated with the center.
By having several spaces, the likelihood of each sample point having at least one appropriate vector space representation is increased.

For a given test sample {\small $\Mat{Z} \in \mathcal{M}$},
the combined output of the {\small $K$} discriminative models can be obtained using:

\vspace{-2ex}
\begin{small}
\begin{equation}
  G(\Mat{Z} | \lambda )
  =
  \sum_{k=1}^K \frac{n_{_k}}{\sum_{l=1}^K{n_{_l}}} ~ g_{_k} \left( \log_{\Mat{\mu}_k}(\Mat{Z}) \right)
  \label{eq:ktangs_predict}
\end{equation}%
\end{small}%

\noindent
where each {\small $n_{_k}$} is used as a mixing coefficient,
emphasising more dense clusters that are less likely to be affected by outliers.

Within the context of pedestrian detection,
the proposed model is trained by first grouping positive samples into {\small $K$} clusters.
Negative samples are not taken into account at this stage as they are not well characterised
(ie.~they do not represent a coherent class).
To obtain the cluster centers, a form of the $k$-means algorithm~\cite{Elkan2003} explicitly adapted to manifolds can be used.
Specifically, 
the geodesic distance in~\eqn{eq:geodesic} is used as the distance measure,
with each cluster center taken to be a Karcher mean, as in~\eqn{eq:mean_def}.
Once the cluster centers are obtained,
all samples are projected onto each tangent space {\small $k$}
and a discriminative model {\small $g_{_k}$} is trained.
The process is summarised in Algorithm~\ref{alg:ktangs}.


The {\it k}-tangent spaces model can be used to replace the weak classifier (which uses a single tangent space)
within the boosting-based algorithm proposed by Tuzel et al.~\cite{TuzelEtAl2008}.
The discriminative models within each {\it k}-tangent spaces model
are the same as the original weak learners (regression functions) used in~\cite{TuzelEtAl2008}.
The resulting strong classifier produced by boosting
is hence comprised of a cascade of {\it k}-tangent spaces.

\begin{algorithm}
  \footnotesize
  \raggedright
  \caption{~Training the {\it k}-tangent spaces model for pedestrian detection}
  \label{alg:ktangs}
  
  \textbf{Input:}
    \begin{itemize}
      \item
      Training samples and their labels:
       $\left\{ (\Mat{X}_i, y_i) \right\}_{i=1}^{N}$, $\Mat{X}_i \in \mathcal{M}, y_i\in\{-1,+1\}$
    \end{itemize}

  \textbf{Processing:}
    \begin{itemize}
      
      \item
      Cluster positive samples to obtain $K$ cluster centers and the number of samples associated with each center:
      $\left\{ (\Mat{\mu}_k, n_{_k}) \right\}_{k=1}^{K}$
      
      \item
      For $k=1, \ldots, K$:
      
      \vspace{-2ex}
      \begin{itemize}
        \item
        Map all samples to the tangent space at $\Mat{\mu}_k$ via $\log_{\Mat{\mu}_k}(\cdot)$
      
        \item
        Train discriminative model $g_k$ using the mapped samples and their labels
        
      \end{itemize}
    \end{itemize}
    
  \vspace{-2ex}
  \textbf{Output:}
    \begin{itemize}
      \item
      Model $\lambda = \left\{ (\Mat{\mu}_k, n_{_k}, g_{_k}) \right\}_{k=1}^{K}$
    \end{itemize}

\end{algorithm}

\begin{figure}[!tb]
  \begin{minipage}{1\columnwidth}
    \begin{minipage}{0.10\textwidth}
      {\small\bf (a)}
    \end{minipage}
    \begin{minipage}{0.8125\textwidth}
      \centering
      \includegraphics[width=0.14\textwidth]{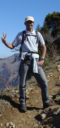}
      \hfill
      \includegraphics[width=0.14\textwidth]{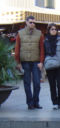}
      \hfill
      \includegraphics[width=0.14\textwidth]{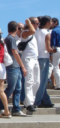}
      \hfill
      \includegraphics[width=0.14\textwidth]{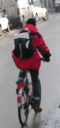}
      \hfill
      \includegraphics[width=0.14\textwidth]{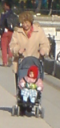}
      \hfill
      \includegraphics[width=0.14\textwidth]{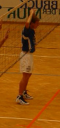}
    \end{minipage}
  \end{minipage}
  
  \vspace{0.5ex}
  
  \begin{minipage}{1\columnwidth}
    \begin{minipage}{0.10\textwidth}
      {\small\bf (b)}
    \end{minipage}
    \begin{minipage}{0.8125\textwidth}
      \centering
      \includegraphics[width=0.14\textwidth]{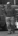}
      \hfill
      \includegraphics[width=0.14\textwidth]{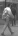}
      \hfill
      \includegraphics[width=0.14\textwidth]{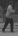}
      \hfill
      \includegraphics[width=0.14\textwidth]{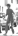}
      \hfill
      \includegraphics[width=0.14\textwidth]{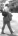}
      \hfill
      \includegraphics[width=0.14\textwidth]{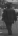}
    \end{minipage}
  \end{minipage}
  \vspace{-2ex}
  \caption
    {
    \small
    Examples of pedestrian images in {\bf (a)}~INRIA and {\bf (b)}~DaimlerChrysler datasets.
    }
  \label{fig:datasets}
  \vspace{-2ex}
\end{figure}

\vspace{-1ex}
\section{Experiments}
\label{sec:experiments}
\vspace{-1.5ex}

We used two datasets of pedestrian images for the experiments:
INRIA~\cite{DalalAndTriggs2005} and DaimlerChrysler~\cite{MunderAndGavrila2006}.
Examples are shown in Fig.~\ref{fig:datasets}.

The INRIA dataset contains 2416 pedestrian samples and 1218 person-free images for training,
as well as 1126 pedestrian samples and 453 person-free images for testing.
The size of the samples is 64$\times$128 with a margin of 16 pixels around the pedestrians.
The main challenges of the INRIA dataset are variations in pose, background and partial occlusions.

The DaimlerChrysler dataset contains three training sets and two tests sets.
Each training and test set contains 4800 pedestrian samples and 5000 negative samples.
Additionally, each training set contains 1200 person-free images where further negative samples can be obtained.
The size of the positive samples is 18$\times$36 with a margin of 2 pixels around the pedestrians.
Detection on this dataset is more difficult due to the small size of the samples
and the challenging negative set.

We performed three experiments:
{\bf (1)}~exploration of the effect of the number of tangent spaces on detection performance;
{\bf (2)}~comparison of the detection performance against other tangent mapping approaches;
{\bf (3)}~comparison of the performance against several notable pedestrian detection methods in the literature.
In all cases the proposed {\it k}-tangent spaces model was used as a weak learner
within the boosting-based pedestrian detection framework proposed by Tuzel et al.~\cite{TuzelEtAl2008}.

As per~\cite{TuzelEtAl2008},
each pixel {\small $I_{(x,y)}$} located at {\small $(x,y)$} was represented by an 8-dimensional feature vector:

\vspace{-2ex}
\begin{footnotesize}
\begin{equation*}
  \Vec{f}_{(x,y)} \mbox{=} \left[ x ~~ y ~~ |d_x| ~~ |d_y| ~~ |d_{xx}| ~~ |d_{yy}| ~~ (d_x^2+d_y^2)^\frac{1}{2} ~ \arctan({|d_x|}/{|d_y|}) \right]
\end{equation*}%
\end{footnotesize}%

\noindent
where
\mbox{\small $d_x = {\partial I_{(x,y)}}/{\partial x} $},
\mbox{\small $d_{xx} = {\partial^2 I_{(x,y)}}/{\partial x^2}$}
with the last two terms representing edge magnitude and orientation.
The covariance descriptor of a given region of pixels (selected by the boosting framework) is hence an 8$\times$8 matrix.

\begin{figure}[!tb]
  \centering
  \vspace{-2ex}
  \includegraphics[width=0.9\columnwidth]{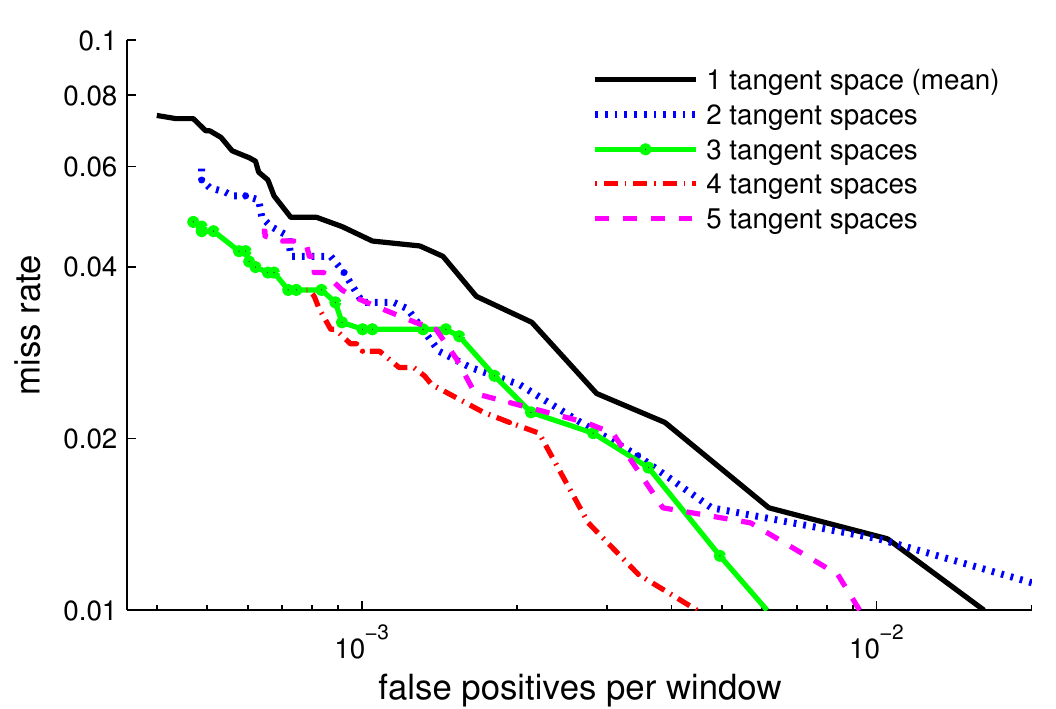}
  \vspace{-2.5ex}
  \caption
    {
    \small
    Effect of the number of tangent spaces on pedestrian detection performance on the INRIA dataset.
    }
  \label{fig:k}
  \vspace{-3ex}
\end{figure}

In the first experiment we varied the number of tangent spaces from 1 to 5.
To avoid training on the background in the INRIA dataset,
the margin around the pedestrians in the positive samples was reduced from 16 to 2 pixels.
The results are presented as detection error trade-off curves in Fig.~\ref{fig:k},
where the best performance point is at the bottom-left corner the graph (ie.~minimal error rates).
The results show that the detection performance generally increases as the number of spaces increases to 4.
Using 5 tangent spaces degrades performance.
We conjecture that the number of training samples affects the optimal number of spaces.
The more spaces are used, the lower the number of samples is available for each cluster,
and hence it is more likely that the clusters are affected by outliers.

\begin{figure}[!tb]
  \begin{minipage}{1\columnwidth}
    \begin{minipage}{0.05\textwidth}
      {\footnotesize\bf (a)}
    \end{minipage}
    \begin{minipage}{0.95\textwidth}
      \includegraphics[width=\textwidth]{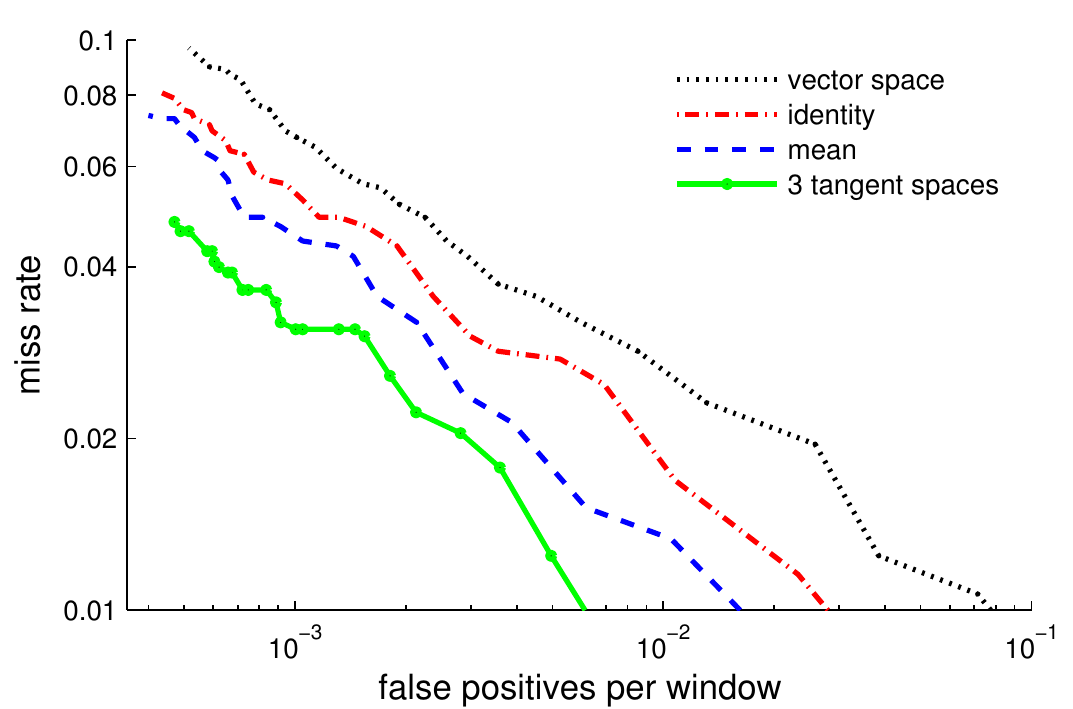}
    \end{minipage}
  \end{minipage}
  
  \begin{minipage}{1\columnwidth}
    \begin{minipage}{0.05\textwidth}
      {\footnotesize\bf (b)}
    \end{minipage}
    \begin{minipage}{0.95\textwidth}
      \includegraphics[width=\textwidth]{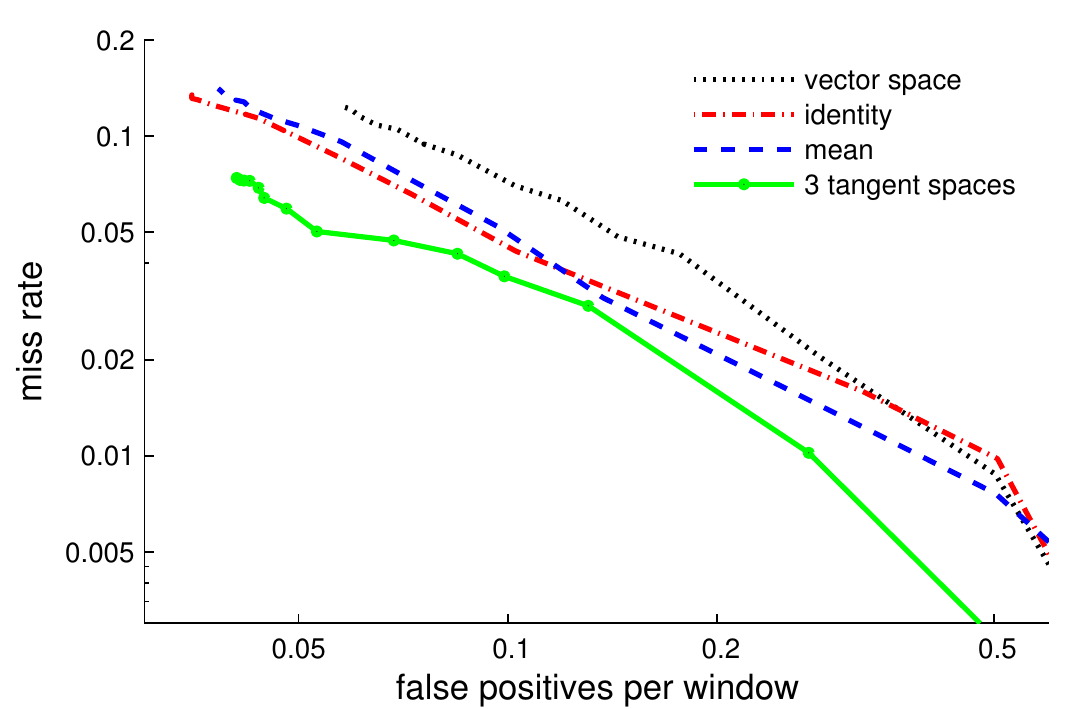}
    \end{minipage}
  \end{minipage}
  
  \vspace{-2ex}
  \caption
    {
    \small
    Pedestrian detection using various tangent space representations on
    {\bf (a)} INRIA
    and
    {\bf (b)} DaimlerChrysler.
    }
  \label{fig:modes}
  \vspace{-2ex}
\end{figure}

In the second experiment, we compared the performance of four tangent mapping approaches:
{\bf (i)}~directly converting the covariance descriptors into vectors (ie.~without projection to a tangent space);
{\bf (ii)}~projecting the points to the tangent space with the identity matrix as the pole;
{\bf (iii)}~projecting the points to the tangent space with the Karcher mean of positive samples as the pole;
{\bf (iv)}~using the proposed \mbox{{\it k}-tangent} spaces model with 3 spaces
(time limitations prevented us from using 4 spaces).

The results, presented in Fig.~\ref{fig:modes},
indicate that any tangent space representation leads to better performance compared to the direct vector space representation.
On the INRIA dataset,
the Karcher mean mapping outperforms the identity matrix mapping,
in agreement with the results presented in~\cite{TuzelEtAl2008}.
However, on the more challenging DaimlerChrysler dataset,
there is little difference between the performance of the Karcher mean and identity matrix mapping.
The proposed {\it k}-tangent spaces model leads to the best performance on both datasets.


In the third experiment we compared the proposed approach
against several notable pedestrian detection methods in the literature.
For the INRIA dataset,
we selected three techniques based on HOG features:
two SVM based methods~\cite{DalalAndTriggs2005},
and an AdaBoost based method~\cite{ZhuEtAl2006}.
For the DaimlerChrysler dataset, 
we selected the best results for the three descriptors proposed in~\cite{MunderAndGavrila2006}:
PCA coefficients, Haar wavelets, and local receptive fields.
For both INRIA and DaimlerChrysler,
we have also compared against the method of Tuzel et al.~\cite{TuzelEtAl2008},
where mapping based on the Karcher mean was used (ie.~single tangent space).

\begin{figure}[!tb]
  \vspace{-0.7ex}
    \begin{minipage}{1\columnwidth}
    \begin{minipage}{0.05\textwidth}
      {\footnotesize\bf (a)}
    \end{minipage}
    \begin{minipage}{0.925\textwidth}
      \includegraphics[width=\textwidth]{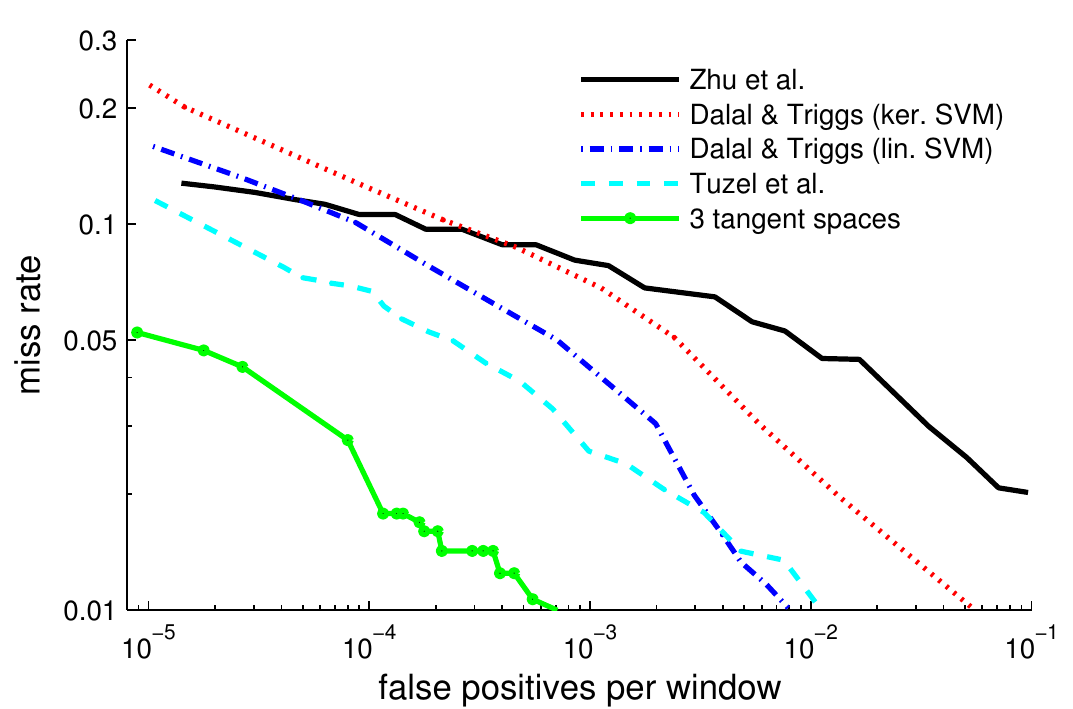}
    \end{minipage}
  \end{minipage}
  
  \vspace{-0.3ex}
  \begin{minipage}{1\columnwidth}
    \begin{minipage}{0.05\textwidth}
      {\footnotesize\bf (b)}
    \end{minipage}
    \begin{minipage}{0.925\textwidth}
      \includegraphics[width=\textwidth]{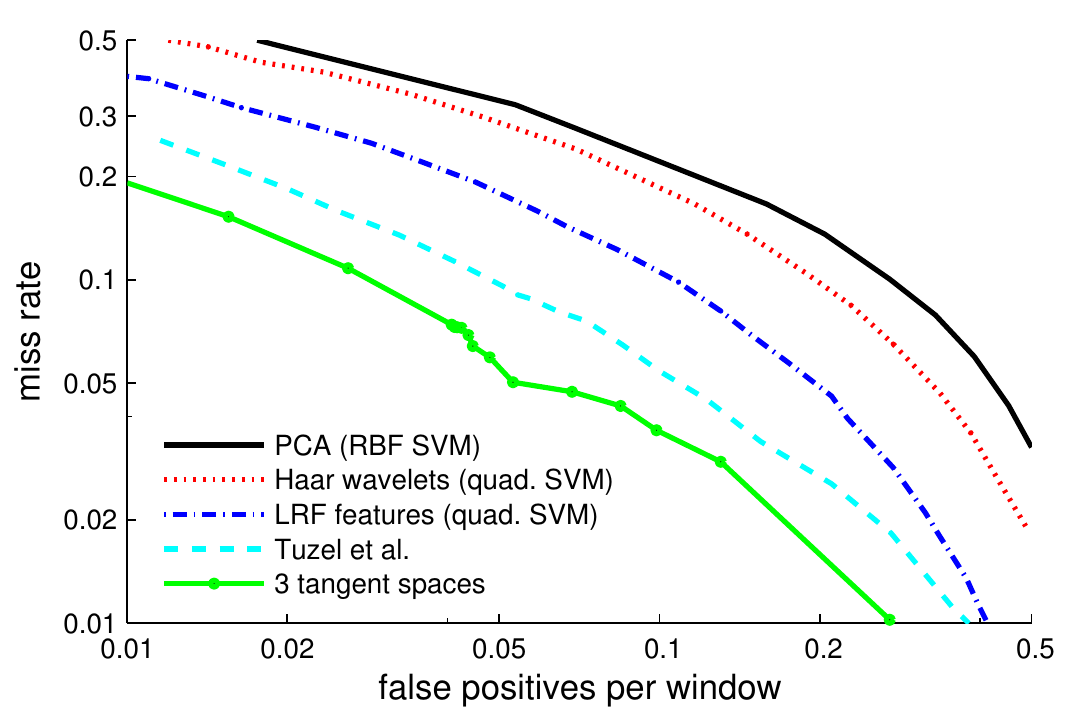}
    \end{minipage}
  \end{minipage}

  \vspace{-1.5ex}
  \caption
    {
    \small
    Comparison with other pedestrian detection methods on
    {\bf (a)}~INRIA
    and
    {\bf (b)}~DaimlerChrysler.
    The configuration of pedestrian images was different than used in Fig.~\ref{fig:modes},
    to allow direct comparison with previously published results
    \mbox{\cite{DalalAndTriggs2005,MunderAndGavrila2006,TuzelEtAl2008,ZhuEtAl2006}}.
    }
  \label{fig:vs}
  \vspace{-3ex}
\end{figure}

To allow a direct comparison with previously published results,
the original images in the INRIA dataset were used (ie.~with a margin of 16 pixels around the pedestrians),
rather the margin reduced images as used in the first and second experiments.
The results,
presented in Fig.~\ref{fig:vs},
indicate that on both datasets considerably better detection performance
is obtained by the proposed {\it k}-tangent spaces approach.

\vspace{-2ex}
\section{Main Findings and Future Directions}
\label{sec:conclusions}
\vspace{-1.5ex}

Covariance descriptors have been recently demonstrated to be useful features for pedestrian detection,
due to their relative low sensitivity to translations and scale variations.
By interpreting the descriptors as points on Riemannian manifolds,
and by exploiting the curvature of such manifolds,
higher performance can be achieved than vector space approaches
such as histograms of oriented gradients.

Current manifold-based classification approaches typically map the points to a tangent space,
where traditional machine learning techniques are used.
However, such treatment is restrictive as distances between arbitrary points on the tangent space
do not represent true geodesic distances, 
and hence do not represent the manifold structure accurately.

In this paper we proposed a general discriminative model based on the combination of several tangent spaces,
in order to preserve more details of the structure.
By having several spaces,
the likelihood of a sample point having at least one appropriate vector space representation is increased.
The model was used as a weak learner within the boosting-based pedestrian detection framework proposed by Tuzel et al.~\cite{TuzelEtAl2008}.
Experiments on the challenging INRIA and DaimlerChrysler datasets
indicate that the proposed model (with just a few tangent spaces)
leads to considerably improved performance.

Future work includes adapting the model to more general classification problems
(ie.~not tied to covariance matrices or pedestrian detection~\cite{Sanderson_AVSS_2012}),
as long as the samples can be represented as points on Riemannian manifolds.


\vspace{-1ex}
\renewcommand{\baselinestretch}{0.855}\small\normalsize
\bibliographystyle{ieee}
\bibliography{references}

\begin{thebibliography}{10}\itemsep=-1pt

\bibitem{DalalAndTriggs2005}
N.~Dalal and B.~Triggs.
\newblock Histograms of oriented gradients for human detection.
\newblock In {\em IEEE Conf. Computer Vision and Pattern Recognition (CVPR)},
  volume~1, pages 886--893, 2005.

\bibitem{DollarEtAl2011}
P.~Dollar, C.~Wojek, B.~Schiele, and P.~Perona.
\newblock Pedestrian detection: an evaluation of the state of the art.
\newblock {\em IEEE Trans. Pattern Analysis and Machine Intelligence},
  34(4):743--761, 2012.

\bibitem{Elkan2003}
C.~Elkan.
\newblock Using the triangle inequality to accelerate k-means.
\newblock In {\em International Conference on Machine Learning}, pages
  147--153, 2003.

\bibitem{FriedmanEtAl2000}
J.~Friedman et~al.
\newblock Additive logistic regression: a statistical view of boosting.
\newblock {\em Annals of Statistics}, 28(2):337--407, 2000.

\bibitem{GualdiEtAl2011}
G.~Gualdi et~al.
\newblock Contextual information and covariance descriptors for people
  surveillance: an application for safety of construction workers.
\newblock {\em EURASIP Journal on Image and Video Processing}, 2011.

\bibitem{GuoEtAl2010}
K.~Guo et~al.
\newblock Action recognition using sparse representation on covariance
  manifolds of optical flow.
\newblock In {\em IEEE Int. Conf. Advanced Video and Signal Based Surveillance
  (AVSS)}, pages 188--195, 2010.

\bibitem{Harandi_ECCV_2012}
M.~T. Harandi, C.~Sanderson, R.~Hartley, and B.~C. Lovell.
\newblock Sparse \mbox{coding} and dictionary learning for symmetric positive
  definite \mbox{matrices}: a kernel approach.
\newblock In {\em European Conference on Computer \mbox{Vision} (ECCV), Lecture
  Notes in Computer Science (LNCS)}, volume 7573, pages 216--229, 2012.

\bibitem{HarandiEtAl2012}
M.~T. Harandi, C.~Sanderson, A.~Wiliem, and B.~C. Lovell.
\newblock Kernel analysis over {R}iemannian manifolds for visual recognition of
  actions, pedestrians and textures.
\newblock In {\em IEEE Workshop on the Applications of Computer Vision}, pages
  433--439, 2012.

\bibitem{Karcher1977}
H.~Karcher.
\newblock {R}iemannian center of mass and mollifier smoothing.
\newblock {\em Comm. Pure and Applied Math.}, 30:509--541, 1977.

\bibitem{MunderAndGavrila2006}
S.~Munder and D.~Gavrila.
\newblock An experimental study on pedestrian classification.
\newblock {\em IEEE Trans. Pattern Analysis and Machine Intelligence},
  28(11):1863--1868, 2006.

\bibitem{PaisitkriangkraiEtAl2008}
S.~Paisitkriangkrai, C.~Shen, and J.~Zhang.
\newblock Fast pedestrian detection using a cascade of boosted covariance
  features.
\newblock {\em IEEE Trans.~Circuits and Systems for Video Technology},
  18(8):1140--1151, 2008.

\bibitem{Sanderson_AVSS_2012}
C.~Sanderson, M.~T. Harandi, Y.~Wong, and B.~C. Lovell.
\newblock Combined learning of salient local descriptors and distance metrics
  for image set face verification.
\newblock In {\em IEEE International Conference on Advanced Video and
  Signal-Based Surveillance (AVSS)}, pages 294--299, 2012.

\bibitem{Shirazi_ICIP_2012}
S.~Shirazi, M.~T. Harandi, C.~Sanderson, and B.~C. Lovell.
\newblock \mbox{Clustering} on {G}rassmann manifolds via kernel embedding with
  application to \mbox{action} analysis.
\newblock In {\em IEEE International Conference on Image \mbox{Processing}},
  pages 781--784, 2012.

\bibitem{TuzelEtAl2008}
O.~Tuzel, F.~Porikli, and P.~Meer.
\newblock Pedestrian detection via classification on {R}iemannian manifolds.
\newblock {\em IEEE Trans. Pattern Analysis and Machine Intelligence},
  30(10):1713--1727, 2008.

\bibitem{ZhuEtAl2006}
Q.~Zhu et~al.
\newblock Fast human detection using a cascade of histograms of oriented
  gradients.
\newblock In {\em IEEE Conf.~Computer Vision and Pattern Recognition}, pages
  1491--1498, 2006.

\end{thebibliography}

\end{document}